\documentclass[letterpaper]{article} 
\usepackage{aaai2026}  
\usepackage{times}  
\usepackage{helvet}  
\usepackage{courier}  
\usepackage[hyphens]{url}  
\usepackage{graphicx} 
\usepackage{natbib}  
\usepackage{caption} 
\usepackage{algorithm}
\usepackage{algorithmic}

\usepackage{tcolorbox}
\usepackage{inconsolata}

\usepackage{microtype}
\usepackage{amsmath}
\usepackage{multirow}
\usepackage{enumitem}
\usepackage{booktabs}

\tcbuselibrary{breakable,skins}
\usepackage{textcomp}

\usepackage{listings}
\usepackage{newfloat}
\usepackage{listings}
\DeclareCaptionStyle{ruled}{labelfont=normalfont,labelsep=colon,strut=off} 
\floatstyle{ruled}
\newfloat{listing}{tb}{lst}{}
\floatname{listing}{Listing}
\newcommand{\ie}{{\it i.e.}}

\newcommand{\data}{PEVData}
\newcommand{\judge}{JELV}

\usepackage{amsmath}
\usepackage{booktabs}
\usepackage{multirow}
\usepackage[table]{xcolor}

\title{JELV: A \textit{J}udge of \textit{E}dit-\textit{L}evel \textit{V}alidity for Evaluation and Automated Reference Expansion in Grammatical Error Correction}

\makeatletter
\def\@fnsymbol#1{$\dagger$}
\makeatother

\author{
    Yuhao Zhan\textsuperscript{\rm 1}, Yuqing Zhang\textsuperscript{\rm 1}, Jing Yuan\textsuperscript{\rm 2}, Qixiang Ma\textsuperscript{\rm 1}, Zhiqi Yang\textsuperscript{\rm 1}, Yu Gu\textsuperscript{\rm 1}, \\ Zemin Liu\textsuperscript{\rm 1}\thanks{Corresponding authors.}, Fei Wu\textsuperscript{\rm 1}\footnotemark[1]}
\affiliations{
    \textsuperscript{\rm 1}Zhejiang University, Hangzhou, China\\
    \textsuperscript{\rm 2}Ludwig-Maximilians-Universität München, Munich, Germany \\

    \{yuhao.zhan, yuqingz7, 3240100569, 3240104460, yuyulo, liu.zemin, wufei\}@zju.edu.cn, jing.yuan@campus.lmu.de
}

\begin{document}

\maketitle

\begin{abstract}
Existing Grammatical Error Correction (GEC) systems suffer from limited reference diversity, leading to underestimated evaluation and restricted model generalization. 
To address this issue, we introduce the \textbf{Judge of Edit-Level Validity (JELV)}, an automated framework to validate correction edits from grammaticality, faithfulness, and fluency. 
Using our proposed human-annotated Pair-wise Edit-level Validity Dataset (PEVData) as benchmark, JELV offers two implementations: a multi-turn LLM-as-Judges pipeline achieving 90\% agreement with human annotators, and a distilled DeBERTa classifier with 85\% precision on valid edits. 
We then apply JELV to reclassify misjudged false positives in evaluation and derive a comprehensive evaluation metric by integrating false positive decoupling and fluency scoring, resulting in state-of-the-art correlation with human judgments.
We also apply JELV to filter LLM-generated correction candidates, expanding the BEA19's single-reference dataset containing 38,692 source sentences. Retraining top GEC systems on this expanded dataset yields measurable performance gains. JELV provides a scalable solution for enhancing reference diversity and strengthening both evaluation and model generalization.
\end{abstract}

\begin{links}
    \link{Code}{https://github.com/yuhao-zhan/JELV}
\end{links}

\section{Introduction}
Grammatical Error Correction (GEC) aims to detect and correct writing errors in text \cite{bryant2023grammatical}. Typical GEC datasets consist of source sentences and their manually corrected versions (\ie, \textit{references}), which form the basis for training and evaluating GEC systems.

\paragraph{Background}
However, creating high-quality corrections for GEC datasets usually requires substantial time and expert effort. Consequently, most GEC datasets \cite{bryant-etal-2019-bea, ng-etal-2014-conll, flachs-etal-2020-grammatical} contain only one or two references per source sentence. This limited reference set does not represent the numerous valid ways an error can be corrected, causing two main issues.
(1) \emph{Underestimated evaluation}: reference-based evaluation metrics undercount acceptable corrections that deviate from the given references \cite{gomez2024multi, zhang2022mucgec, choshen-abend-2018-inherent}. 
(2) \emph{Limited model performance}: training on a narrow reference set constrains the model to specific correction patterns, leading to poor generalization.
While prior studies show that increasing the number of references improves both evaluation accuracy \cite{bryant2015far} and training effectiveness \cite{liu2024towards}, existing reference expansions are still created manually. This manual approach hinders scalability, restricts the diversity of valid corrections, and perpetuates inherent biases in reference-based evaluation \cite{choshen-abend-2018-inherent}. Thus, an automated approach for expanding diverse and valid corrections is crucial to advance research and development in GEC.

\begin{figure}
  \includegraphics[width=\columnwidth]{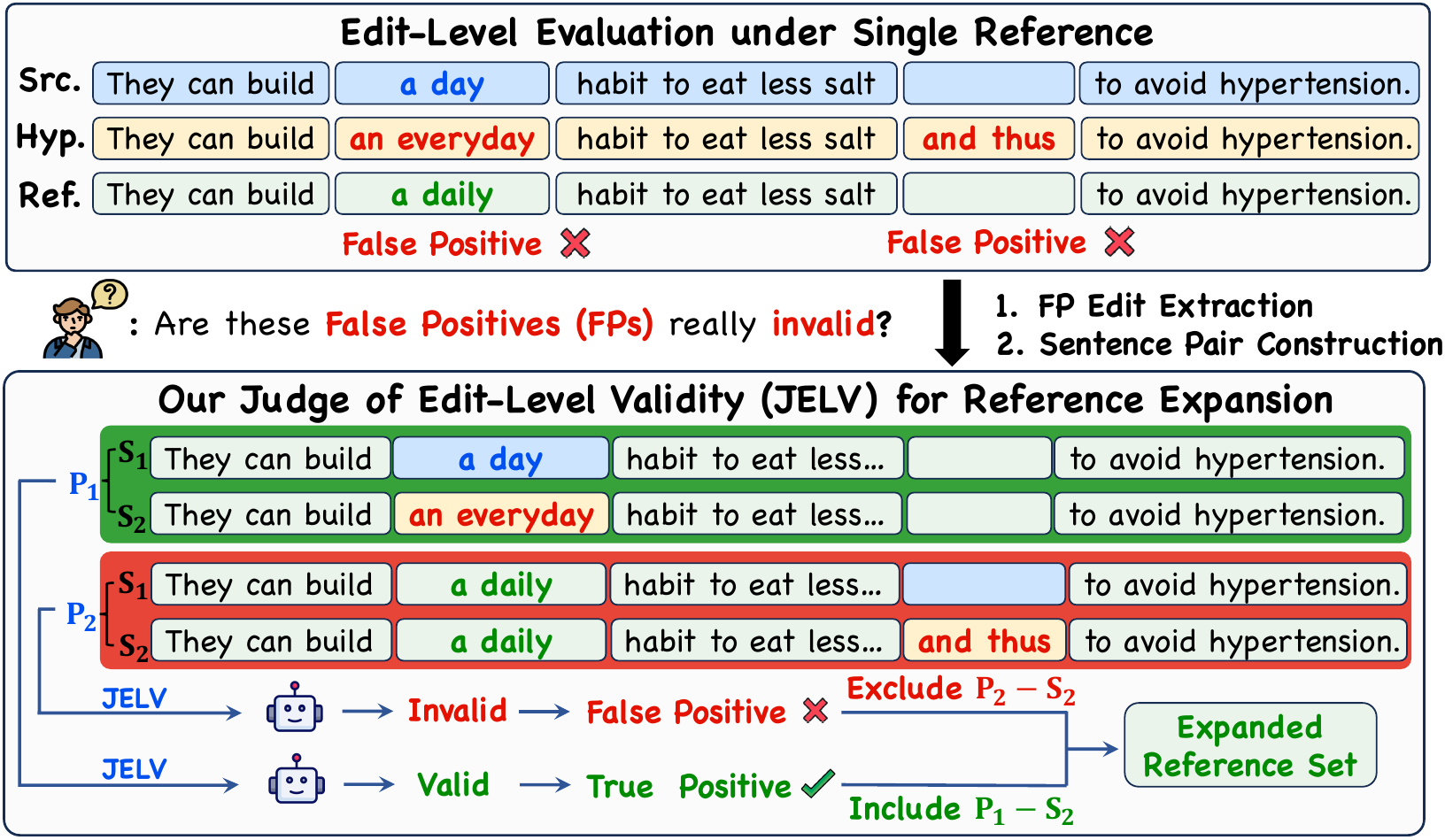}
  \caption{Comparison between biased edit‐level evaluation using a single reference and our automated expansion of valid edits via the Judge of Edit‐Level Validity (JELV). \textbf{Src.}, \textbf{Hyp.} and \textbf{Ref.} denote the source, hypothesis and reference sentences, respectively. We form two sentence pairs, $\mathbf{P_1}$ and $\mathbf{P_2}$. In each pair, $\mathbf{S_1}$ (from \textbf{Src.}) and $\mathbf{S_2}$ (from \textbf{Hyp.}) differ only in the single edited segment, remaining identical to the reference elsewhere.}
  \label{fig:first_image}
\end{figure}

However, clearly defining what makes a correction edit \textit{valid} remains challenging in GEC. Current human annotation practices follow two primary guidelines: \emph{Minimal Edit} enforces grammaticality and faithfulness\footnote{Faithfulness: Corrections should maintain the original textual meaning and syntactic structure.} \cite{ye2024cleme2}, and \emph{Fluent Edit} further requires improvements in sentence fluency \cite{bryant2023grammatical}. 

While most existing datasets adopt the Minimal Edit approach, a more comprehensive standard for validity should recognize overall improvements to the text, including fluency. Based on this, we define an edit as \textbf{valid} only if it satisfies all three golden criteria: (1) grammatical correctness, (2) meaning preservation, and (3) fluency improvement.

\paragraph{Present work}
To provide a gold‐standard benchmark for edit‐level validity, we introduce the Pair‐wise Edit‐level Validity Dataset (PEVData). 
To collect valid edits that are often overlooked, we notice that many actually \textit{valid} hypothesis edits\footnote{candidate correction produced by the GEC system} are misclassified as \textit{invalid}---False Positive (FP)---in reference-based evaluation due to incomplete reference coverage, as shown in Figure \ref{fig:first_image}. Therefore, we extract every hypothesis sentence containing edits initially classified as FPs and pair it with its source sentence. We then align all other tokens to the reference so that only single edit span differs. Expert annotators subsequently evaluate each pair's validity against our three golden criteria, producing the \data\footnote{For more details on \data, please refer to Sec.~\ref{sec.data}.}.

With \data \ as benchmark, we introduce the \textbf{J}udge of \textbf{E}dit‐\textbf{L}evel \textbf{V}alidity (\textbf{\judge}) to automate edit validity assessment. \judge\ offers two implementations to balance accuracy and inference efficiency. JELV1.0 leverages the evaluation capacity of large language models (LLMs) \cite{li2024llms, zhu2023judgelm} and implements a multi-turn LLM-as-Judges pipeline, achieving over 90\% accuracy compared to human labels on \data. To reduce inference cost and enable large-scale deployment, we distill this pipeline into JELV2.0, a lightweight DeBERTa \cite{he2020deberta} classifier that maintains over 85\% precision on valid edits. With JELV’s high‐precision edit judgments, valid corrections are automatically identified and integrated into the reference set. The complete automated reference expansion workflow is shown in Figure \ref{fig:first_image}.

We apply \judge\ to mitigate reference scarcity in two complementary ways. (1) \textbf{Implicit} reference expansion for \textit{underestimated evaluation}. It is infeasible to eliminate misclassified FPs through exhaustive explicit reference expansion since even short sentences have over 1000 valid corrections on average \cite{choshen-abend-2018-inherent}. Instead, we introduce \judge\ into evaluation pipeline to reclassify those misclassified FPs as true positives (TPs)—achieving the equivalent of exhaustive reference enumeration at minimal cost. By further decoupling the remaining FPs into overcorrection and non-overcorrection and integrating fluency scoring for fine-grained and comprehensive evaluation, we achieve state-of-the-art correlation with human judgments across multiple evaluation dimensions. (2) \textbf{Explicit} reference expansion for \textit{limited model performance}. We employ a generation‐then‐filtering pipeline to automate explicit reference expansion. Leveraging LLMs’ GEC expertise and lower cost than human annotation, we generate candidate edits for BEA19’s \cite{bryant-etal-2019-bea} 38,692 single-reference sentences and use JELV to retain only valid ones. Retraining top GEC systems on this expanded corpus yields clear performance gains on CoNLL14 benchmark. This demonstrates the effectiveness of our reference expansion strategy in improving model performance.

\paragraph{Contributions}
Our contributions are threefold.
(1) We introduce \judge\ to automate edit-level validity assessment, offering an LLM-as-Judges pipeline with $>$90\% accuracy and a distilled DeBERTa classifier with $>$85\% precision, both validated on our human-annotated \data\ benchmark.
(2) We enhance evaluation reliability by using \judge\ to reclassify FPs as TPs, decoupling the remaining FPs, and integrating fluency scoring, achieving SOTA correlation with human judgments.
(3) We propose a \judge-based generation-then-filtering pipeline for automated reference expansion and retraining top GEC systems on the expanded corpus yields measurable performance gains. 


\section{Related Work}
We only discuss the most relevant studies here and provide further discussion in Appendix A.

\paragraph{GEC Evaluation}
Automatic evaluation metrics for GEC are divided into \textit{reference-based} and \textit{reference-less} approaches \cite{maeda-etal-2022-impara}. \textbf{Reference-based metrics} score outputs by comparing them to human references. These methods can penalize valid corrections that are not in the reference set, reflecting biases in limited reference coverage \cite{choshen-abend-2018-inherent}. \textbf{Reference-less metrics} \cite{yoshimura-etal-2020-reference, maeda-etal-2022-impara} assess correction quality without references using statistical and language models, but may lack transparency and inherit biases from their underlying models \cite{deutsch-etal-2022-limitations}. \textbf{LLM-based metrics} \cite{kobayashi-etal-2024-large,xie2024dsgram} use LLMs as scorers to evaluate corrections, offering strong human correlation at the expense of higher computational cost and potential instability. \textbf{Meta-evaluation} benchmarks such as GJG15 \cite{grundkiewicz-etal-2015-human} and SEEDA \cite{kobayashi2024revisiting} measure the correlation between GEC metrics and human judgments, providing the standard for validating GEC evaluation methods. However, GJG15 was later found to be problematic, casting doubt on many conclusions based on it \cite{choshen-abend-2018-automatic, chollampatt-ng-2018-reassessment}. SEEDA, on the other hand, performs human evaluations based on two different granularities: edit- (SEEDA-E) and sentence-level (SEEDA-S), making it more reliable.


\section{Validity Judgment}
This section establishes our framework for automated edit-level validity judgment. We first define clear criteria for valid edits, then introduce our human-annotated benchmark dataset \data, and finally present \judge\ 1.0 and 2.0.

\subsection{Criteria for Validity}
We begin by analyzing edits (\ie, \textit{source}$\rightarrow$ \textit{target}) that are truly \textbf{valid} but misclassified as FPs due to incomplete reference coverage. In GEC, these FPs fall into two categories: (1) \textbf{FP$\mathbf{_{oc}}$} (overcorrection): valid edits applied to grammatically correct text for improving fluency or semantic clarity. (2) \textbf{FP$\mathbf{_{noc}}$} (non-overcorrection): valid edits applied to ungrammatical text but absent from reference set. 

\begin{figure}[t]
  \centering
  \includegraphics[width=0.75\columnwidth]{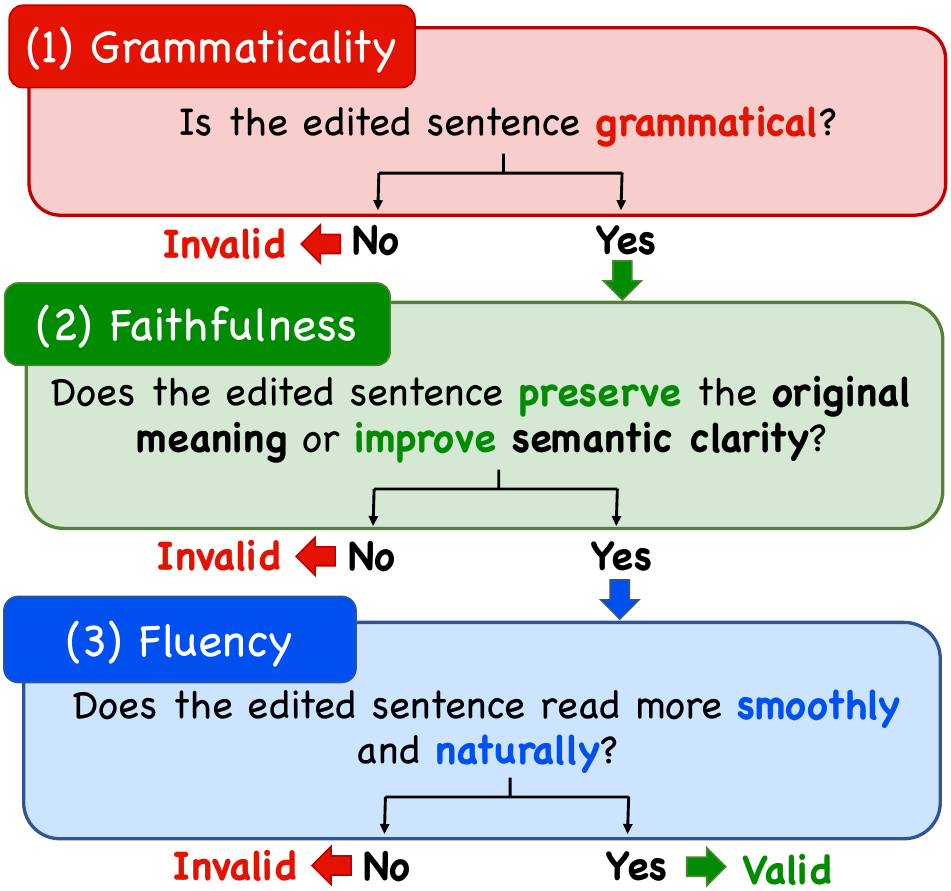}

  \caption{Three criteria for judging edit-level validity.}
  \label{fig:criteria}

\end{figure}

For each reference, under the assumption that all text outside the edit span is correct, we define three criteria for edit-level validity to ensure comprehensive improvement: \textbf{(1) Grammaticality}: the edit target must be free of grammatical errors. \textbf{(2) Faithfulness}: the edit must preserve the original intended meaning or enhance semantic clarity. \textbf{(3) Fluency}: the edit must render the sentence more fluent, either by correcting errors or smoothing awkward phrasing.
An edit is judged valid only if it meets all three criteria simultaneously (see Figure \ref{fig:criteria}).

\subsection{\data: Dataset Curation} \label{sec.data}
To construct a benchmark for edit-level validity, we assemble Pair-wise Edit-level Validity Dataset (\data) in following three stages.

\paragraph{Sentence Pair Construction}
Under the Correction Independence Assumption that grammatical error corrections in a sentence are independent
\cite{ye-etal-2023-cleme}, we extract each hypothesis sentence containing edits initially judged as FPs and pair it with its source. All other tokens are realigned to the reference so that only single edit span differs. This controlled pairing isolates each edit’s effect and prevents unrelated changes from influencing validity judgments. \label{construction protocol}

\begin{figure*}[t]
\centering
\includegraphics[width=0.75\linewidth]{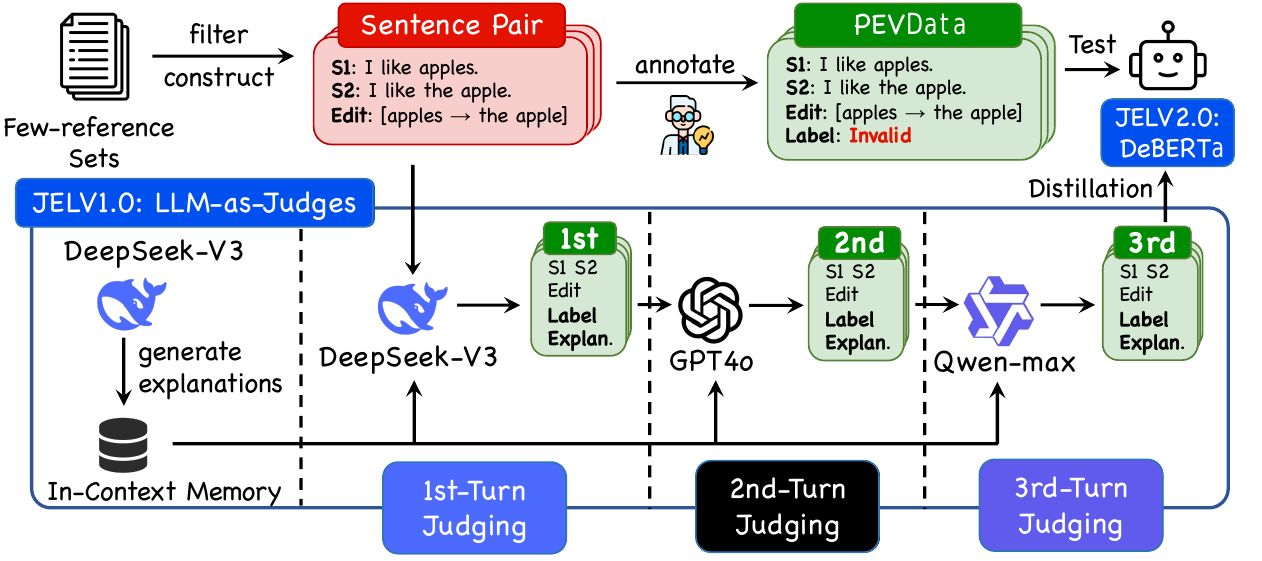}

  \caption{Overview of the \judge\ workflow. Starting from few reference sets, we extract candidate sentence pairs and process them in two independent streams. One stream is manually annotated by experts to create the \data. The other is evaluated by a three turn LLM as Judges pipeline (\judge1.0) and the resulting labels are distilled into a DeBERTa classifier (\judge2.0).}
  \label{fig:overview_1}
\end{figure*}

\paragraph{Data Collection}
We sample single‐edit pairs from three public datasets: 1,118 from \emph{CoNLL14} \cite{ng-etal-2014-conll}, 844 from \emph{ArgRewrite} \cite{zhang-etal-2017-corpus}, and 835 from \emph{JFLEG} \cite{napoles-etal-2017-jfleg}. We provide more details of these datasets in Appendix B.

\paragraph{Annotation Protocol}
University experts (in English teaching, proofreading, or linguistics) judged all pairs for edit validity, viewing each with preceding and following sentences for context. For \emph{CoNLL14} (no prior validity labels), three annotators applied a three-stage process: (1) an independent pass, retaining unanimous decisions; (2) a second independent pass on remaining pairs, requiring unanimity; and (3) a joint discussion. This process yielded IAA (Fleiss' Kappa) 0.81, with remaining conflicts resolved by majority vote. By contrast, for \emph{ArgRewrite} pairs, which already carried ``Better/NotBetter'' judgments from seven prior annotators, a single expert performed a final validity check. \emph{JFLEG} pairs, drawn from the reference set, were accepted as valid without annotation. In total, \data\ comprises 2,797 pairs (1,459 valid, 1,338 invalid), forming a balanced benchmark; its error diversity also matches \emph{CoNLL14} (Simpson Diversity: 0.912 vs. 0.914).

\subsection{\judge: Developing an Automated Judge}

While \data\ provides manually annotated judgments on edit-level validity, automatic evaluation at scale is essential. Therefore, we propose the Judge of Edit-Level Validity (\judge) and demonstrate its effectiveness on \data\ benchmark.
Besides, to balance accuracy and efficiency, \judge\ comprises two implementations: \emph{\judge1.0}, an LLM-as-judges pipeline for high accuracy, and \emph{\judge2.0}, a distilled DeBERTa classifier for fast inference. Figure \ref{fig:overview_1} shows the \judge\ workflow overview.

\subsubsection{\judge1.0: LLM-as-Judges Pipeline}

\begin{itemize}
    \item \textbf{Explain-then-Annotate.} Before giving a judgment, the model generates a one-sentence explanation for its decision to ensure transparency and interpretability.
    \item \textbf{In-Context Learning (ICL).} 
    To increase \judge1.0's accuracy, we help it understand our criteria through ICL. We use DeepSeek-V3 (DS-V3) to generate explanations for 50 edits sampled from \data\footnote{We limit to 50 edits (25 valid, 25 invalid) to preserve test set integrity while ensuring balanced representation.} (excluded from the test set) against our three validity criteria, creating an \textit{in-context memory}. For each judgment, we include one valid and one invalid example as two-shot demonstrations.
    

    \item \textbf{Context‐Aware Prompting.}  We include the preceding and following sentences of the sentence being judged
    as additional context for more informed judgments. 
    \item \textbf{Multi-Turn Optimization.} \judge1.0 sequentially applies DS-V3, GPT-4o, and Qwen-Max. Each model reviews and refines the previous model's explanations and judgments, correcting errors through iterative calibration. 
    We use Qwen-Max’s outputs as the final validity labels.
    See Appendix G.1 for prompt details.
\end{itemize}

\begin{table}
  \centering
  \resizebox{0.4\textwidth}{!}{
  \begin{tabular}{lcccc}
    \hline
    \textbf{LLM} & \textbf{Prec.} & \textbf{Rec.} & \textbf{$\mathbf{F_{0.5}}$} & \textbf{Accuracy}\\
    \hline
    DS-V3 & 0.6920 & \textbf{0.9038} & 0.7261 & 0.8415 \\
    GPT4o & 0.8417 & 0.8349 & 0.8403 & 0.8976\\ 
    Qwen& \textbf{0.8771} & 0.8462 & \textbf{0.8707} & \textbf{0.9134}
     \\\hline
  \end{tabular}
  }
  \caption{Comparison of LLM-as-Judges \judge1.0 predictions and human annotations on the \data\ benchmark. \textbf{Prec.} and \textbf{Rec.} denote precision and recall, respectively. \textbf{Bold} indicates the \textbf{highest} score.}
  \label{tab:LLM-as-Judges}
\end{table}

\begin{figure}[ht]
\centering
  \includegraphics[width=0.8\columnwidth]
  {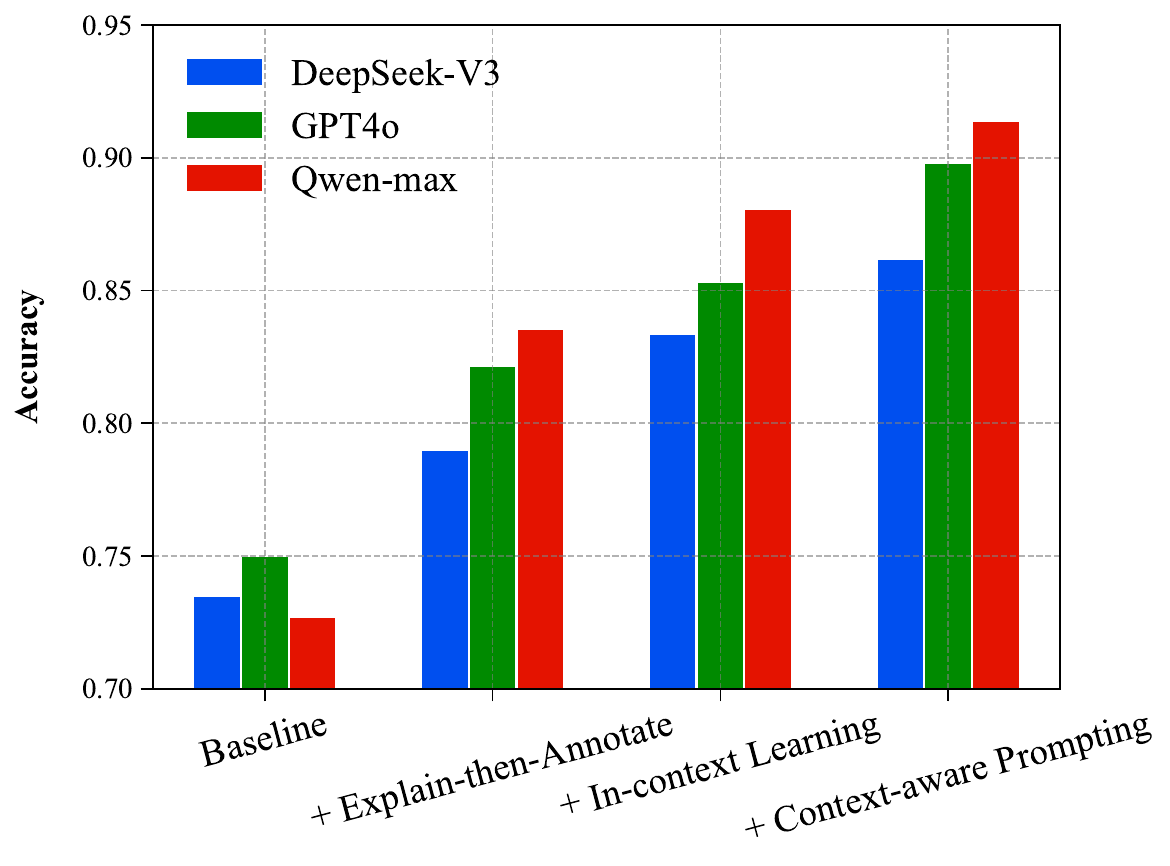}

  \caption{Ablation of \judge1.0 strategies on \data.}
  \label{fig:ablation_JLEV1.0}

\end{figure}

Table \ref{tab:LLM-as-Judges} reports how well the LLM-as-Judges predictions align with human annotations on \data\ benchmark. Higher values indicate better alignment with human judgments. We use the F$_{0.5}$, which weights precision 2$\times$ recall, to prioritize correct identification of \textit{valid} edits for reference quality. The third-turn Qwen-max achieves the highest F$_{0.5}$ (0.8707) and Accuracy (0.9134), demonstrating our pipeline's effectiveness. \footnote{The $\sim$10\% inconsistency is mainly due to highly subtle fluency edits (e.g., help build → foster), cases so nuanced that even experts require discussion to reach consensus.} Figure \ref{fig:ablation_JLEV1.0} presents an ablation study evaluating the contribution of each component in \judge1.0 on \data\ benchmark. Starting from the baseline of independent LLM judgments, we find that adding ``explain-then-annotate'', then ICL, and finally context-aware prompting each delivers a clear accuracy increase. Moreover, every subsequent LLM outperforms its predecessor, highlighting the strength of multi-turn optimization. When all three strategies are combined, overall accuracy exceeds 90\%.

\subsubsection{\judge2.0: Classifier Model Distillation}

While \judge1.0 achieves high alignment with human judgments, it incurs high inference costs that limit large-scale deployment. To address this, we distill \judge1.0 into a lightweight DeBERTa classifier ($<$1B parameters). Using our sentence pair construction approach (Section \ref{construction protocol}), we extract candidate edits from the extended CoNLL14 reference sets \cite{bryant2015far}, yielding 12,984 sentence pairs. After removing overlaps with \data\ to avoid data leakage, 12,416 pairs remain. We then apply \judge1.0 to label these pairs, producing 5,786 valid and 6,630 invalid examples as training data. We train the DeBERTa classifier on this dataset, reserving half of \data\ for validation and half for testing.  

To enhance \judge2.0's judgment accuracy, we integrate two auxiliary features that capture complementary aspects of sentence pairs: GPT-2 probability (indicating grammaticality and fluency \cite{yasunaga-etal-2021-lm}) and SBERT semantic similarity (reflecting meaning preservation)\footnote{See ablation study on the two features in Appendix C.}. These features are projected and concatenated with DeBERTa's final embeddings via a lightweight classification head to avoid overwhelming the primary text representations. To address class imbalance, we apply focal loss \cite{lin2017focal}. We enhance robustness through FreeLB adversarial training \cite{zhu2019freelb}. Our training employs curriculum learning \cite{bengio2009curriculum} with gradual layer unfreezing to prevent overfitting and the final model is selected using stratified k-fold cross validation with early stopping.
On the test set, \judge2.0 achieves 85.25\% precision, an F$_{0.5}$ score of 82.24\%, and 78.91\% accuracy, demonstrating its high precision with low inference latency.


\begin{figure*}[t]
\centering
\includegraphics[width=0.75\linewidth]{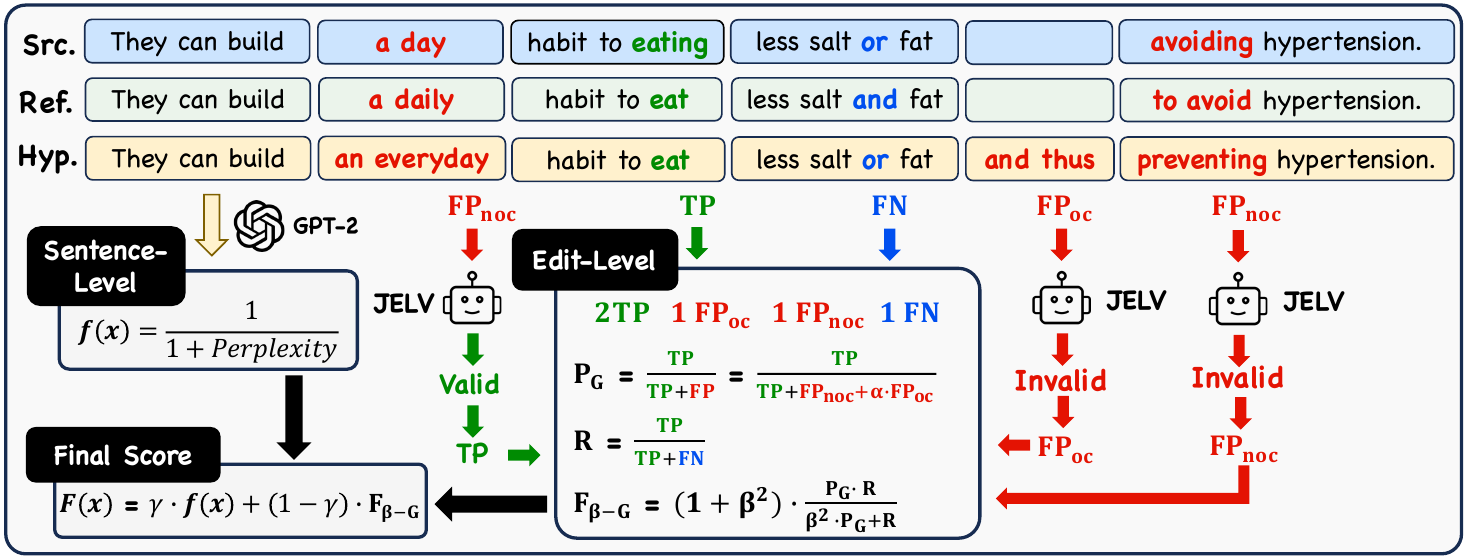}

\caption{Overview of the comprehensive evaluation metric. Edits flagged as FPs are first reclassified via JELV2.0, then false positive decoupling distinguishes the remaining $\mathrm{FP_{{noc}}}$ and $\mathrm{FP_{{oc}}}$ to compute the edit‐level generalized F‐score $\mathrm{F_{\beta\text{-}G}}$. In parallel, GPT-2 perplexity produces the sentence‐level fluency score $f(x)$ for the hypothesis. A interpolation weight $\gamma$ combines these two streams into the final metric $F(x)$.}
  \label{fig:EvaluationMetric}
\end{figure*}

\section{Evaluation}

This section presents how we integrate \judge2.0 into GEC evaluation pipelines to derive a comprehensive metric with additional evaluation strategies, and demonstrate its effectiveness through meta-evaluation.
\subsection{Method}

\paragraph{JELV-based Reclassification}  
\cite{choshen-abend-2018-inherent} shows that even short sentences have over a thousand valid corrections on average,  making exhaustive reference expansion infeasible and causing reference-based metrics to misclassify some valid edits as FPs inevitably. To mitigate this inherent bias, we apply JELV2.0 to re-evaluate each edit initially marked as a FP during evaluation. Edits judged valid by JELV2.0 are reclassified as TPs, achieving
the equivalent of exhaustive inclusion of JELV-validated references in evaluation (\ie, implicit reference expansion) without incurring the cost of explicit expansion, which reduces misclassification and enhances evaluation reliability. After reclassifying edits, we apply CLEME‐independent \cite{ye-etal-2023-cleme}, a leading F-score based metric, to compute F$_{0.5}$ scores and this yields the \textit{JELV‐based CLEME}.  

\paragraph{False Positive Decoupling}  
FPs in GEC fall into two categories: non-overcorrections (FP$\mathrm{_{noc}}$), which edit ungrammatical text, and overcorrections (FP$\mathrm{_{oc}}$): which alter already correct text. Considering the two different edit properties that deserve different penalties, we apply a relative penalty weight \(\alpha\) to FP$\mathrm{_{oc}}$ (\ie, overcorrection). We then derive the \textit{generalized precision}:
\begin{equation}\label{generalized_precision}
\mathrm{P_{G}} 
  = \mathrm{\frac{TP}{TP + FP}} = \mathrm{\frac{TP}{TP + FP_{noc} + \alpha\,FP_{oc}}}.
\end{equation}
Combining \(\mathrm{P_{G}}\) with recall \(\mathrm{R}\), we further propose the \textit{generalized F-score} as the edit-level metric:
\begin{equation}\label{generalized_F}
    \mathrm{F_{\beta-{G}}} 
      = (1+\beta^2)\cdot \mathrm{\frac{P_G\cdot R}{\beta^2\cdot P_G + R}}.
\end{equation}

\paragraph{Fluency Score Integration}  
Prior study \cite{kobayashi2024revisiting} recommends the combination of both edit-level and sentence-level metrics in GEC evaluation. Therefore, we incorporate a sentence-level fluency evaluation based on GPT-2 perplexity \cite{ge2018reaching}:
\begin{equation}
\label{fluency_eval}
    f(x) = \frac{1}{1+H(x)}, \ H(x) = -\frac{\sum_{i=1}^{|x|}logP(x_i|x_{<i})}{|x|},
\end{equation}
where \(P(x_{i} | x_{<i})\) is computed by GPT-2 and \(|x|\) denotes sentence length. Higher \(f(x)\) indicates greater fluency.

\paragraph{Comprehensive Metric}  
Building on the complementary strengths of edit-level and sentence-level evaluations, we combine \(\mathrm{F_{\beta\text{-}G}}\) and the fluency score \(f(x)\) via an interpolation weight \(\gamma\):
\begin{equation}\label{comprehensive_metric}
F(x) \;=\;(1 - \gamma)\,\cdot F_{\beta\text{-}G} \;+\;\gamma\cdot\,f(x)\,.
\end{equation}
Figure \ref{fig:EvaluationMetric} presents an overview of our comprehensive evaluation metric.

\begin{table*}[ht]
\centering
\resizebox{\textwidth}{!}{
\begin{tabular}{l|cc|cc|cc|cc|cc|cc|cc|cc}
\hline

 & \multicolumn{8}{c|}{\textbf{System-level}} & \multicolumn{8}{c}{\textbf{Sentence-level}} \\
 \multirow{2}{*}{\textbf{Metric}}

 & \multicolumn{4}{c|}{\textbf{SEEDA-E}} & \multicolumn{4}{c|}{\textbf{SEEDA-S}} & \multicolumn{4}{c|}{\textbf{SEEDA-E}} & \multicolumn{4}{c}{\textbf{SEEDA-S}} \\

  & \multicolumn{2}{c|} {Base} & \multicolumn{2}{c|} {+ Fluent corr.} & \multicolumn{2}{c|} {Base} & \multicolumn{2}{c|} {+ Fluent corr.} & \multicolumn{2}{c|} {Base} & \multicolumn{2}{c|} {+ Fluent corr.} & \multicolumn{2}{c|} {Base} & \multicolumn{2}{c} {+ Fluent corr.} \\

 & $r$ & $\rho$ &$r$&$\rho$&$r$&$\rho$&$r$&$\rho$&$Acc$& $\tau$ &$Acc$& $\tau$ &$Acc$& $\tau$ &$Acc$& $\tau$ \\
\hline
M$^2$ \cite{dahlmeier-ng-2012-better}                          & 0.711  & 0.699   & -0.224 & 0.160   & 0.654  & 0.462   & -0.293 & 0.002   & 0.578  & 0.156   & 0.458  & -0.084  & 0.567  & 0.135   & 0.454  & -0.092  \\
    ERRANT \cite{bryant-etal-2017-automatic}                   & 0.628  & 0.629   & -0.531 & 0.024   & 0.516  & 0.364   & -0.604 & -0.143  & 0.511  & 0.215   & 0.460  & 0.118   & 0.485  & 0.174   & 0.438  & 0.085   \\
    CLEME \cite{ye-etal-2023-cleme}             & 0.644  & 0.587   & -0.480 & 0.051   & 0.547  & 0.336   & -0.552 & -0.108  & 0.443  & 0.188   & 0.398  & 0.078   & 0.428  & 0.175   & 0.390  & 0.077   \\
    CLEME2.0 \cite{ye-etal-2025-cleme2}             & 0.836  & 0.888   & -0.589 & 0.187   & 0.716  & 0.636   & -0.665 & 0.029   & 0.364  & 0.182   & 0.364  & 0.182   & 0.583  & 0.583   & 0.583  & 0.583   \\
    PT-M$^2$ \cite{gong-etal-2022-revisiting}                       & 0.820  & 0.902   & -0.187 & 0.305   & 0.820  & 0.720   & -0.244 & 0.191   & 0.220  & -0.031  & 0.220  & -0.028  & 0.228  & -0.003  & 0.218  & -0.024  \\
    GLEU \cite{napoles-etal-2015-ground}         & 0.886  & 0.867   & 0.206  & 0.600   & 0.820  & 0.790   & 0.118  & 0.547   & 0.673  & 0.365   & 0.616  & 0.247   & 0.671  & 0.358   & 0.610  & 0.234   \\
    SOME \cite{yoshimura-etal-2020-reference}                     & 0.860  & 0.923   & 0.926  & 0.952   & 0.868  & 0.776   & 0.917  & 0.859   & 0.394  & -0.034  & 0.405  & -0.027  & 0.409  & -0.014  & 0.405  & -0.036  \\
    Scribendi Score \cite{islam-magnani-2021-end}        & 0.718  & 0.692   & 0.696  & 0.745   & 0.513  & 0.413   & 0.592  & 0.547   & 0.377  & 0.241   & 0.357  & 0.208   & 0.369  & 0.247   & 0.324  & 0.176   \\
    IMPARA \cite{maeda-etal-2022-impara}                & 0.843  & 0.937   & 0.860  & 0.960   & 0.878  & \textbf{0.860}   & 0.853  & \textbf{0.912} & 0.695  & 0.413   & 0.700  & 0.414   & 0.721  & 0.473   & 0.713  & 0.445   \\
    LLM-based      & 0.951  & 0.977   & 0.917  & 0.981   & 0.919  & 0.858 & 0.881  & 0.906   & 0.717  & 0.536   & 0.692  & 0.500   & 0.753  & 0.614   & 0.711  & \textbf{0.557} \\
    JELV-based $\mathrm{F(x)}$                  & \textbf{0.975} & \textbf{0.986} & \textbf{0.974} & \textbf{0.991} & \textbf{0.932} & \textbf{0.860} & \textbf{0.947} & \textbf{0.912} & \textbf{0.780} & \textbf{0.559} & \textbf{0.770} & \textbf{0.541} & \textbf{0.807} & \textbf{0.630} & \textbf{0.772} & 0.543   \\

\hline
\end{tabular}
}
\caption{Results of system- and sentence-level meta-evaluations on SEEDA's test set of social media domain. For ``LLM-based'', our prompt and experiment setting
aligns with previous works’s protocol \cite{kobayashi-etal-2024-large}. \textbf{Boldface} indicates the \textbf{highest} correlation score in each column.}

\label{main_table: eval}

\end{table*}

\subsection{Experiments}
\label{experiment}

We conduct comprehensive experiments to evaluate our metric against existing approaches. Our evaluation consists of hyperparameter tuning on a training set, followed by meta-evaluations on a held-out test set of SEEDA benchmark.


\paragraph{Settings}
To reach higher correlation with human judgments at each level, we tune the hyperparameters $\alpha$ and $\gamma$ on the training set by selecting the values that maximize the correlation\footnote{$\alpha \in [0, 2],\ \gamma \in [0,1]$, both with step size 0.01.}. Once identified, these optimal parameters are fixed and applied to the metric for evaluation on the test set. To prevent any in-domain overlap, we exploit SEEDA’s two distinct domains—genetic (163 source sentences) and social media (228 source sentences)—by assigning one domain entirely to training and the other entirely to testing. Finally, we assess our metric’s robustness via system-level and sentence-level meta-evaluations on the test set and compare its performance against existing metrics.

\paragraph{Meta‐evaluation}
SEEDA provides edit‐level (SEEDA‐E) and sentence‐level (SEEDA‐S) pairwise human judgments for twelve GEC system corrections (``Base'') and two additional fluent human corrections (``+Fluent corr.'')
For \textbf{system‐level evaluation}, we compute the overall scores across all corrections for each GEC system, derive a system ranking, and compare it to the human ranking produced by TrueSkill \cite{sakaguchi-etal-2014-efficient} using Pearson’s \(r\) and Spearman’s \(\rho\). For \textbf{sentence‐level evaluation}, we compute a score for each correction sentence, then for every pair of corrections we derive a metric-based preference, compare this to SEEDA’s human judgment, and report classification accuracy and Kendall’s \(\tau\).

\subsection{Result and Analysis}

\paragraph{Main Results}

Table \ref{main_table: eval} reports our meta-evaluation on SEEDA’s test set of social media domain, where our metric achieves state-of-the-art (SOTA) correlations with human judgments across nearly all evaluation levels and granularities. Results for the test set of genetic domain are provided in the Appendix D, showing similarly strong performance and confirming our metric’s robustness across domains.

\begin{figure}[t]
    \centering
    \includegraphics[width=\linewidth]{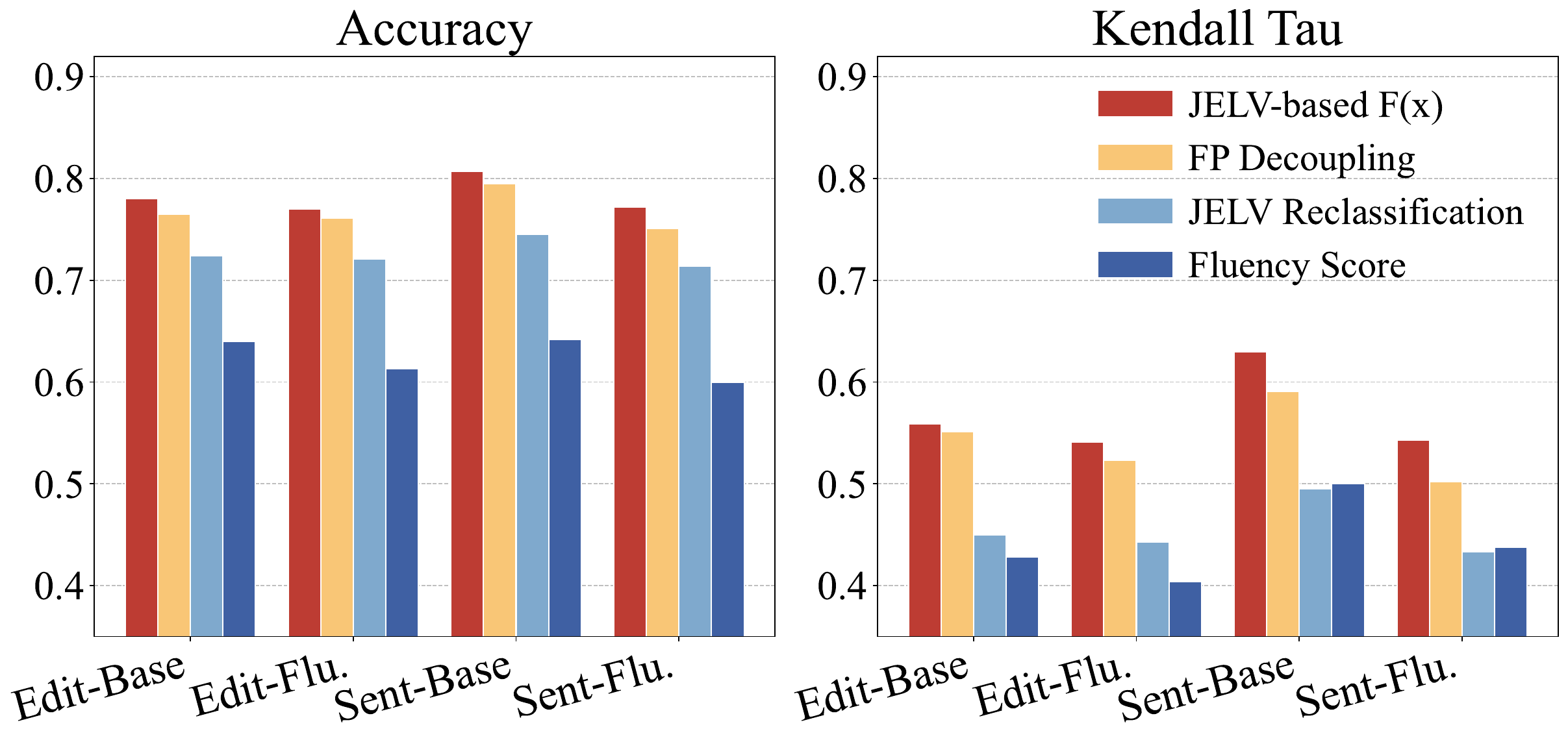}
    \caption{Ablation Study for JELV-based F(x).}
    \label{fig: ablation_eval}
\end{figure}

\begin{figure}[ht]
    \centering
    \includegraphics[width=0.6\linewidth]{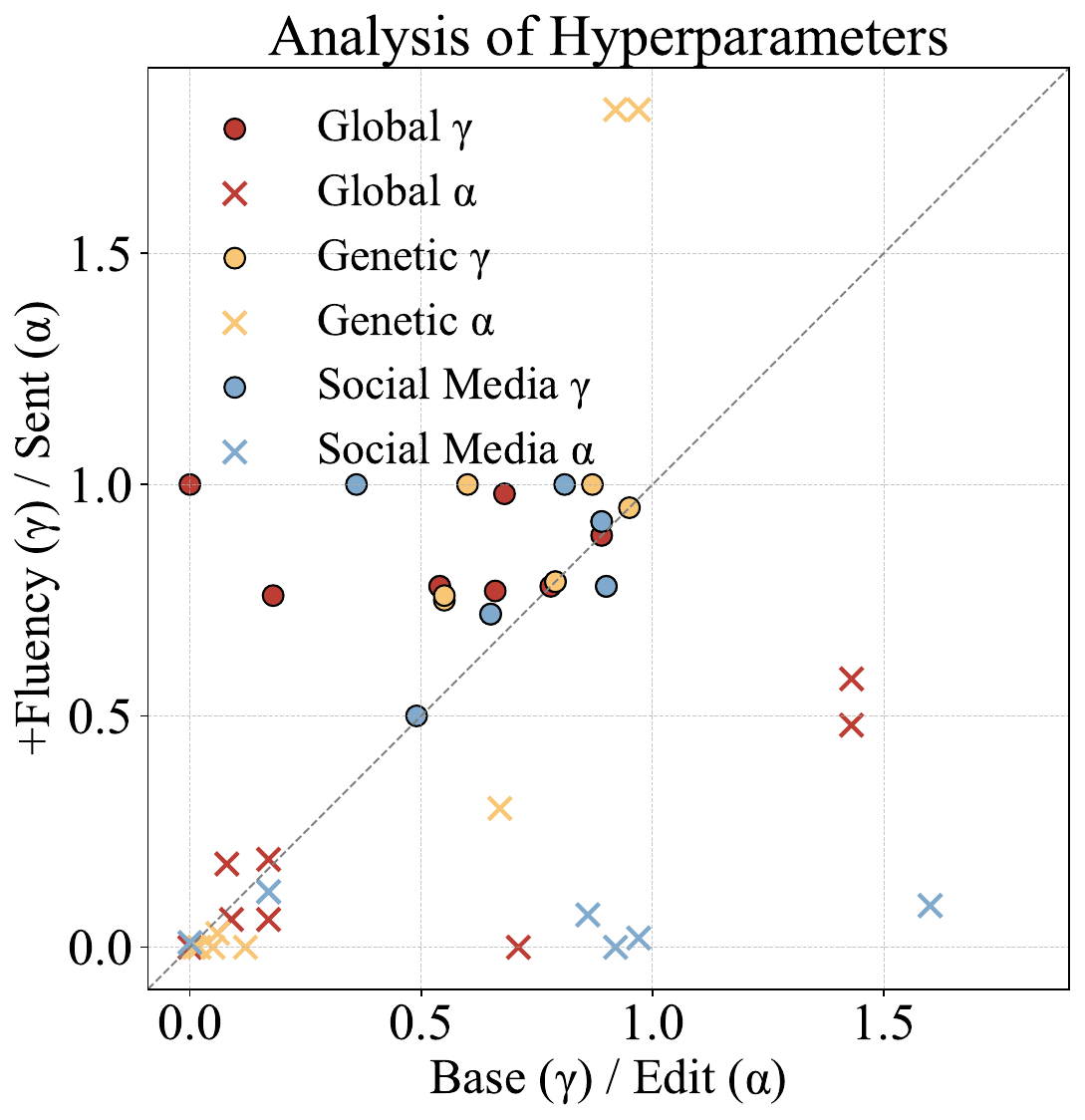}
    \caption{Comparison of optimal hyperparameters across SEEDA domains and evaluation levels. Base($\gamma$) and +Fluency($\gamma$) denote optimal $\gamma$ for ``Base'' and ``+Fluent corr.'' settings, respectively; Edit($\alpha$) and Sent($\alpha$) denote optimal $\alpha$ for edit- and sentence-level evaluation, respectively.}
    \label{fig:alpha and gamma}
\end{figure}

\paragraph{Ablation Study}
We perform ablation studies to isolate each component—disabling JELV-based reclassification, fixing $\alpha$=1 to remove FP decoupling, and fixing $\gamma$=0 to exclude fluency score integration. Figure \ref{fig: ablation_eval}'s sentence-level meta-evaluation results show that omitting any component reduces correlation with human judgments and thus confirms the necessity of each method in the final metric.

\begin{figure*}[ht]
  \centering
  \includegraphics[width=0.75\linewidth]{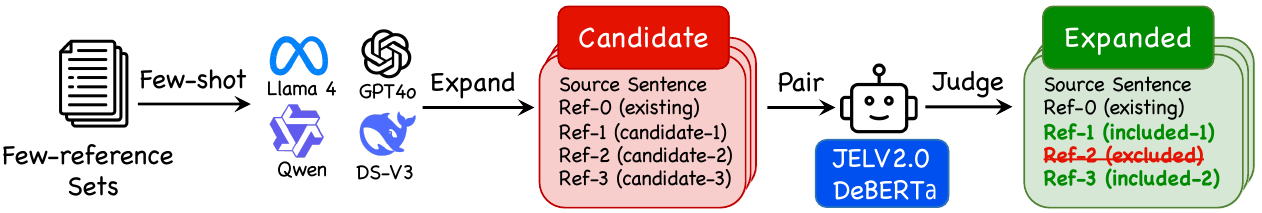}

  \caption{Automated expansion pipeline for large scale few reference sets. Four LLMs generate candidate corrections using few shot prompts. Each candidate correction is paired with its source sentence following the Sentence Pair Construction protocol and evaluated by the distilled DeBERTa classifier (\judge 2.0). Valid corrections are added to the reference set to produce the expanded corpus.}
  \label{fig:overview_2}

\end{figure*}

\begin{table*}[t]
  \centering
  \resizebox{\textwidth}{!}{
  \begin{tabular}{l*{9}{c}}
    \toprule
    \textbf{Dataset}
      & \textbf{Ensembles} & \textbf{Majority-voting} & \textbf{GRECO} & \textbf{GEC-DI}& \textbf{UGEC}& \textbf{MoECE}& \textbf{SynGEC}& \textbf{Sequence tagging}\\
    \midrule
    Raw      & 72.68	&71.44	&70.99	&69.43	&69.38	&	67.45&	67.29&	66.31   \\ 
    Expanded w/o Filtering     & 72.98 &	71.44	 &71.25 &	69.57	 &69.62		 &67.74	 &67.47	 &66.36    \\
    \rowcolor{gray!20}
    Expanded w/ Filtering      & \textbf{73.05}   & \textbf{71.91}    & \textbf{71.36}    & \textbf{69.58}  & \textbf{69.74}  & \textbf{67.92}  & \textbf{67.57}  & \textbf{66.67}    \\

    \midrule
  Raw & 73.54 & 71.84 & 71.34 & 69.37 & 69.44 & 66.87 & 66.68 & 65.09 \\
Expanded w/o Filtering & 74.08 & 72.25 & 71.60 & \textbf{70.02} & 70.00 & 67.13 & 66.75 & 65.23 \\
\rowcolor{gray!20} Expanded w/ Filtering & \textbf{74.19} & \textbf{72.87} & \textbf{72.16} & 69.84 & \textbf{70.14} & \textbf{67.59} & \textbf{67.16} & \textbf{65.96} \\

    \midrule

    Raw & 79.47 & 78.10 & 78.08 & 77.34 & 77.80 & 76.78 & 76.80 & 75.42 \\
Expanded w/o Filtering & 80.29 & 79.60 & 78.98 & \textbf{78.51} & 78.14 & 76.47 & 76.26 & 75.70 \\
\rowcolor{gray!20} Expanded w/ Filtering & \textbf{80.45} & \textbf{80.01} & \textbf{79.57} & 78.41 & \textbf{78.85} & \textbf{77.35} & \textbf{77.33} & \textbf{76.95} \\

    \bottomrule
  \end{tabular}
  }

 \caption{Evaluation scores for eight GEC systems on CoNLL14 under three BEA19 training variants. Every score is averaged over the twice identical retraining runs to reduce bias.  Each group of bars (top to bottom) shows M$^2$, CLEME, and JELV-based $\mathrm{F(x)}$ scores. Raw: reproduction trained on raw BEA19; Expanded w/o Filtering: BEA19 plus all LLM-generated candidates; Expanded w/ Filtering: BEA19 plus JELV-validated edits. \textbf{Bold} indicates the highest score.}

  \label{table: result of BEA19 expanded}

\end{table*}

\paragraph{Analysis of Hyperparameter}
We investigate how the overcorrection penalty \(\alpha\) and fluency weight \(\gamma\) vary across evaluation settings and granularities. We expect edit-level evaluation to require a larger \(\alpha\) (Equation \ref{generalized_precision}) than sentence-level evaluation, because it penalizes individual corrections more strictly. 
We also expect the ``+Fluent corr.'' setting to require a larger \(\gamma\) (Equation \ref{comprehensive_metric}) than the ``Base'' setting, since it places greater emphasis on fluency. To verify this, we extract the optimal \(\alpha\) and \(\gamma\) values for the genetic, social media, and overall SEEDA domains at each evaluation level. Figure \ref{fig:alpha and gamma} shows that most of the edit-level \(\alpha\) exceeds the sentence-level \(\alpha\), and that the ``+Fluent corr.'' \(\gamma\) exceeds the ``Base'' \(\gamma\), confirming our hypotheses and demonstrating the flexibility and robustness of our hyperparameter design. All the optimal hyperparameters can be found in Appendix E.

\section{Reference Expansion and Retraining}

Figure~\ref{fig:overview_2} shows our automated two-stage pipeline for expanding references: generation followed by filtering. We apply this pipeline to the BEA-2019 train and dev sets \cite{bryant-etal-2019-bea}, which comprise comprising 38,692 source sentences paired with a single human reference. 

\paragraph{Generation}
To leverage LLMs’ GEC expertise and reduce reliance on human, we use four LLMs (Llama 4, GPT-4o, Qwen and DS-V3) to generate correction candidates. Each LLM is prompted with the source sentence and its single reference as a one-shot example, and is briefed on our three validity criteria to guide output corrections toward high-quality valid edits (see full prompt in Appendix G.2). On average, these LLMs produce 8.74 (Llama 4), 6.82 (GPT-4o), 4.06 (Qwen) and 2.41 (DS-V3) candidates per sentence.

\paragraph{Filtering}
Each candidate edit is evaluated by JELV2.0, and only edits judged valid are included in the final expanded reference set. JELV-based filtering on BEA-train reduces the average corrections per sentence from 11.54 to 3.90 and the maximum from 104 to 40 (see Appendix F for full statistics). This demonstrates JELV’s effectiveness in filtering diverse LLM-generated corrections into a concise, high-quality reference set.

\paragraph{Retraining}
To assess the impact of our reference expansion strategy on model performance, we retrain eight top GEC systems\footnote{Systems are selected from the CoNLL14 leaderboard, and training follows the original papers’ experiment protocols.} twice on three BEA19 variants: the raw train and dev sets (baseline); the expanded set (BEA19 plus LLM-generated candidates); and the filtered set (expanded set with JELV validation). The systems include Ensembles, Majority-voting \cite{omelianchuk2024pillars}, GRECO \cite{qorib2023system}, GEC-DI \cite{zhou2023improving}, Unsupervised GEC \cite{cao2023unsupervised}, MoECE \cite{qorib2024efficient}, SynGEC \cite{zhang2022syngec}, and GECTOR \cite{omelianchuk2020gector}. Table~\ref{table: result of BEA19 expanded} presents each system’s performance under three training variants, evaluated with M$^2$, CLEME, and JELV-based $\mathrm{F(x)}$. Training on the expanded and filtered dataset yields the highest performance for most systems—significantly outperforming the Raw baseline (paired t-test and Wilcoxon signed-rank test: both p $<$ 0.01), especially on JELV-based $\mathrm{F(x)}$—Unfiltered expansion also improves over the baseline for most systems, while JELV filtering delivers further gains, underscoring the value of JELV-based quality-controlled reference expansion. While the modest gains are expected from highly optimized SOTA baselines, our strategy's key effect is enabling models to learn new correction patterns for some rare but challenging errors they typically fail on. Achieving consistent improvements on such errors thus confirms the value of our method.

\section{Conclusions}
In this paper, we introduce Judge of Edit-Level Validity (\judge), a framework for automated edit validity assessment in GEC. \judge\ offers two implementations: \judge1.0, a multi-turn LLM-as-judges pipeline, and \judge2.0, a distilled DeBERTa classifier. Both achieve strong agreement with human annotations on our \data.
We demonstrate two applications that mitigate reference scarcity. 
First, our evaluation metric applies \judge\ to reclassify false positives and then incorporates false positive decoupling together with fluency scoring, resulting in state-of-the-art correlation with human judgments.
Second,  a \judge-based generation-then-filtering pipeline automates reference expansion and retraining on the expanded dataset yields clear improvements in GEC model performance. 
\judge\ therefore provides a scalable solution to enrich reference diversity and enhance both evaluation reliability and model robustness in GEC.

\section*{Ethical Statement}
Our study builds on publicly available GEC datasets. Three expert annotators with bac kgrounds in English teaching, proofreading, and linguistics manually judged edit validity and were compensated at market rates. We thank them for their contributions. To our knowledge, there are no ethical concerns arising from the use of these data or annotations.  

\section*{Acknowledgements}
We sincerely thank the anonymous reviewers for their insightful comments and constructive feedback, which significantly improved the quality of this paper.

\section*{Appendix}

\setcounter{section}{0}
\setcounter{subsection}{0}

\setcounter{secnumdepth}{5}
\renewcommand{\thesection}{\Alph{section}}
\renewcommand{\thefigure}{\Roman{figure}}
\renewcommand{\thetable}{\Roman{table}}

\section{Detailed Related Work}
\label{app: related_work}
\subsection{GEC Reference Expansion}
Most GEC training corpora, including BEA19-train \cite{bryant-etal-2019-bea}, provide only one reference per source sentence, while evaluation benchmarks typically include two \cite{bryant-etal-2019-bea, ng-etal-2014-conll, flachs-etal-2020-grammatical} or four \cite{napoles-etal-2017-jfleg} references. Manual reference expansion has improved evaluation reliability by adding eight annotations for CoNLL14 \cite{bryant2015far}, averaging 2.3 references per sentence in the Chinese MuCGEC dataset \cite{zhang2022mucgec}, and providing three references per sentence in a Russian GEC corpus \cite{gomez2024multi}. Besides, \cite{liu2024towards} has shown that multiple references have great potential for better training effectiveness. However, these approaches depend on extensive human effort and do not scale easily. An automated, scalable method for generating diverse, valid correction edits is therefore essential for advancing GEC.

\subsection{GEC Evaluation Metrics}
A core component of a GEC system is the ability to measure model performance. \textbf{Reference-based metrics}, the traditional paradigm, rely on alignment with human-crafted references. Early approaches like the M$^2$ scorer used edit-based F$_{0.5}$ scoring but faced criticism for artificially inflated precision \cite{bryant-etal-2017-automatic}, prompting linguistically grounded alternatives like ERRANT \cite{bryant-etal-2017-automatic} for improved edit alignment. Despite their effectiveness, reference-based metrics often suffer from issues such as overfitting to reference-specific patterns and inevitably penalizing valid but non-reference corrections, which introduces inherent biases \cite{choshen-abend-2018-inherent}. To circumvent reliance on references, \textbf{reference-less metrics} emerged, leveraging quality estimation frameworks: Grammaticality-Based Metrics \cite{napoles-etal-2016-theres} combined fluency and meaning preservation scores, later extended by SOME \cite{yoshimura-etal-2020-reference} and IMPARA \cite{maeda-etal-2022-impara} with BERT-based models and Scribendi Score \cite{islam-magnani-2021-end} using perplexity-edit hybrids. However, these metrics, while bypassing the need for references, can lack interpretability and may inherit biases from the models they employ \cite{deutsch-etal-2022-limitations}, and struggle to detect issues like over-correction. \textbf{The advent of LLMs} introduced a paradigm shift, with GPT-4 \cite{kobayashi-etal-2024-large} and Prompt Engineering \cite{xie2024dsgram} enabling direct numerical scoring via natural language criteria interpretation, aiming to mimic human-like evaluation and potentially capturing more complex aspects of correction quality. Despite their promise, LLM-based metrics face challenges in explainability (acting as ``black-box'' evaluators), instability, multiple biases \cite{li2024llms}, and prohibitive computational costs. 

\subsection{Meta-evaluation} 
Meta-evaluation frameworks determine metric validity: human correlation. This involves evaluating metrics themselves by correlating their scores with human judgments based on two dominant datasets GJG15 \cite{grundkiewicz-etal-2015-human} and SEEDA \cite{kobayashi2024revisiting}, providing a gold standard for determining the true effectiveness of GEC evaluation methodologies and guiding the development of superior metrics. However, GJG15 was later found to be problematic, and many of the conclusions drawn using these datasets were called into question\cite{choshen-abend-2018-automatic, chollampatt-ng-2018-reassessment}. Additionally, GJG15 may yield different results depending on granularity, such as sentence-level or edit-level evaluations\cite{kobayashi2024revisiting}. SEEDA, on the other hand, performs human evaluations based on two different granularities: SEEDA-E for edit-based evaluation and SEEDA-S for sentence-based evaluation, making it more reliable. 

\begin{table*}[ht]
\centering
\resizebox{\textwidth}{!}{
\begin{tabular}{l|cc|cc|cc|cc|cc|cc|cc|cc}
\hline

 & \multicolumn{8}{c|}{\textbf{System-level}} & \multicolumn{8}{c}{\textbf{Sentence-level}} \\
 \multirow{2}{*}{\textbf{Metric}}

 & \multicolumn{4}{c|}{\textbf{SEEDA-E}} & \multicolumn{4}{c|}{\textbf{SEEDA-S}} & \multicolumn{4}{c|}{\textbf{SEEDA-E}} & \multicolumn{4}{c}{\textbf{SEEDA-S}} \\

  & \multicolumn{2}{c|} {Base} & \multicolumn{2}{c|} {+ Fluent corr.} & \multicolumn{2}{c|} {Base} & \multicolumn{2}{c|} {+ Fluent corr.} & \multicolumn{2}{c|} {Base} & \multicolumn{2}{c|} {+ Fluent corr.} & \multicolumn{2}{c|} {Base} & \multicolumn{2}{c} {+ Fluent corr.} \\

 & $r$ & $\rho$ &$r$&$\rho$&$r$&$\rho$&$r$&$\rho$&$Acc$& $\tau$ &$Acc$& $\tau$ &$Acc$& $\tau$ &$Acc$& $\tau$ \\
\hline
    M$^2$                 & 0.806  & 0.841   & 0.049   & 0.429   & 0.658   & 0.637   & -0.062  & 0.266   & 0.602  & 0.205   & 0.543   & 0.087   & 0.563   & 0.127   & 0.511   & 0.022   \\
    ERRANT               & 0.732  & 0.748   & -0.436  & 0.099   & 0.558   & 0.448   & -0.534  & -0.090  & 0.559  & 0.253   & 0.516   & 0.174   & 0.546   & 0.245   & 0.477   & 0.114   \\
    CLEME                & 0.766  & 0.825   & -0.395  & 0.240   & 0.653   & 0.587   & -0.483  & 0.081   & 0.514  & 0.266   & 0.468   & 0.164   & 0.488   & 0.222   & 0.429   & 0.091   \\
    CLEME2.0             & 0.808  & 0.790   & -0.576  & 0.125   & 0.679   & 0.566   & -0.656  & -0.015  & 0.545  & 0.364   & 0.545   & 0.364   & 0.615   & 0.615   & 0.615   & \textbf{0.615}   \\
    PT-M$^2$             & 0.838  & 0.874   & 0.036   & 0.363   & 0.743   & 0.734   & -0.057  & 0.257   & 0.230  & -0.018  & 0.227   & -0.023  & 0.231   & 0.011   & 0.224   & 0.001   \\
    GLEU                 & 0.886  & 0.867   & 0.206   & 0.600   & 0.820   & 0.790   & 0.118   & 0.547   & 0.675  & 0.390   & 0.666   & 0.362   & 0.641   & 0.320   & 0.613   & 0.253   \\
    SOME                 & 0.912  & 0.923   & 0.943   & 0.952   & 0.880   & \textbf{0.881} & 0.929   & \textbf{0.925} & 0.432  & 0.019   & 0.434   & 0.017   & 0.396   & -0.045  & 0.403   & -0.036  \\
    Scribendi Score      & 0.858  & 0.818   & 0.638   & 0.763   & 0.687   & 0.699   & 0.532   & 0.695   & 0.449  & 0.316   & 0.442   & 0.310   & 0.404   & 0.243   & 0.386   & 0.213   \\
    IMPARA               & 0.931  & 0.923   & 0.921   & 0.952   & \textbf{0.917} & \textbf{0.881} & 0.902   & \textbf{0.925} & 0.773  & 0.555   & 0.653   & 0.312   & 0.769   & 0.542   & 0.642   & 0.286   \\
    LLM-based            & \textbf{0.959} & 0.923   & 0.898   & 0.947   & 0.906   & 0.846   & 0.851   & 0.899   & 0.724  & 0.571   & 0.720   & 0.577   & 0.709   & 0.547   & 0.692   & 0.531 \\
    JELV-based $\mathrm{F(x)}$.           & 0.956 & \textbf{0.986} & \textbf{0.974} & \textbf{0.991} & 0.883  & 0.846   & \textbf{0.947} & 0.899   & \textbf{0.827} & \textbf{0.654} & \textbf{0.826} & \textbf{0.649} & \textbf{0.792} & \textbf{0.585} & \textbf{0.763} & 0.526  \\

\hline
\end{tabular}
}
\caption{Results of system- and sentence-level meta-evaluations on SEEDA's test set of genetic domain. For ``LLM-based'', our prompt and experiment setting
aligns with previous works’s protocol \cite{kobayashi-etal-2024-large}. \textbf{Boldface} indicates the \textbf{highest} correlation score in each column.}

\label{main_table: eval_another}

\end{table*}

\begin{table*}[ht]
\centering
\resizebox{\textwidth}{!}{
\begin{tabular}{l|cc|cc|cc|cc|cc|cc|cc|cc}
\hline
 & \multicolumn{8}{c|}{\textbf{System-level}} & \multicolumn{8}{c}{\textbf{Sentence-level}} \\
 \multirow{2}{*}{\textbf{Metric}}

 & \multicolumn{4}{c|}{\textbf{SEEDA-E}} & \multicolumn{4}{c|}{\textbf{SEEDA-S}} & \multicolumn{4}{c|}{\textbf{SEEDA-E}} & \multicolumn{4}{c}{\textbf{SEEDA-S}} \\

  & \multicolumn{2}{c|} {Base} & \multicolumn{2}{c|} {+ Fluent corr.} & \multicolumn{2}{c|} {Base} & \multicolumn{2}{c|} {+ Fluent corr.} & \multicolumn{2}{c|} {Base} & \multicolumn{2}{c|} {+ Fluent corr.} & \multicolumn{2}{c|} {Base} & \multicolumn{2}{c} {+ Fluent corr.} \\

 & $r$ & $\rho$ &$r$&$\rho$&$r$&$\rho$&$r$&$\rho$&$Acc$& $\tau$ &$Acc$& $\tau$ &$Acc$& $\tau$ &$Acc$& $\tau$ \\
    \midrule
    $\alpha$ (Global)       & 0.09 & 0.17 & 0.08 & 0.17 & 0.06 & 0.06 & 0.18 & 0.19 
                            & 0.71 & 1.43 & 0.00 & 1.43 & 0.00 & 0.48 & 0.00 & 0.58 \\
    $\alpha$ (Genetic)      & 0.67 & 0.92 & 0.12 & 0.97 & 0.30 & 1.81 & 0.00 & 1.81 
                            & 0.02 & 0.06 & 0.05 & 0.01 & 0.00 & 0.03 & 0.00 & 0.00 \\
    $\alpha$ (Social Media) & 0.86 & 0.92 & 0.00 & 0.97 & 0.07 & 0.00 & 0.01 & 0.02 
                            & 1.60 & 1.60 & 0.17 & 0.17 & 0.09 & 0.09 & 0.12 & 0.12 \\
    \midrule
    $\gamma$ (Global)       & 0.66 & 0.78 & 0.77 & 0.78 & 0.54 & 0.54 & 0.78 & 0.78 
                            & 0.68 & 0.89 & 0.98 & 0.89 & 0.00 & 0.18 & 1.00 & 0.76 \\
    $\gamma$ (Genetic)      & 0.87 & 0.89 & 1.00 & 0.92 & 0.60 & 0.95 & 1.00 & 0.95 
                            & 0.79 & 0.79 & 0.79 & 0.79 & 0.55 & 0.55 & 0.75 & 0.76 \\
    $\gamma$ (Social Media) & 0.81 & 0.89 & 1.00 & 0.92 & 0.36 & 0.49 & 1.00 & 0.50 
                            & 0.90 & 0.90 & 0.78 & 0.78 & 0.65 & 0.65 & 0.72 & 0.72 \\
    \bottomrule
  \end{tabular}
  }
  \caption{Tuned optimal hyperparameters $\alpha$ and $\gamma$ for different evaluation settings across domains.}
  \label{tab:hyperparams}
\end{table*}

\section{Dataset Collection} \label{app:data-collect}

We sample single‐edit pairs from three public datasets: 
\begin{itemize}[topsep=0pt,itemsep=-1ex,partopsep=1ex,parsep=1ex,leftmargin=*]
  \item \textbf{CoNLL14.} \cite{ng-etal-2014-conll,bryant2015far} the official test set provides two references per sentence, with an eight-reference extension offering additional corrections. Noting that some of these extended edits are not so reliable, we extract 1,118 edits unique to this extension for our expert annotation on validity.
  \item \textbf{ArgRewrite.} \cite{zhang-etal-2017-corpus} Seven prior annotators labeled each revision as ``Better'' or ``NotBetter''. However, “Better” does not guarantee that an edit satisfies all three of our validity criteria. We therefore select 844 single-edit revisions with at least five ``Better'' votes for our expert validity re-annotation.
  \item \textbf{JFLEG.} \cite{napoles-etal-2017-jfleg} Designed for \textit{fluency} evaluation, this dataset provides four references per source sentence. We extract 835 single-edit references, pair each with its source sentence, and treat them as valid since they originate from existing references.
\end{itemize}

\section{Ablation Study for JELV2.0's Features}
\begin{table}[ht]
  \centering
  \begin{tabular}{lccc}
    \toprule
    \textbf{Model}            & \textbf{Precision} & \boldmath$F_{0.5}$\textbf{ (\%)} & \textbf{Accuracy} \\
    \midrule
    Baseline                  & 80.48\%                 & 79.33\%                         & 72.59\%                \\
    + GPT-2                    & 83.01\%                 & 81.67\%                         & 75.29\%                \\
    + SBERT                    & 82.44\%                 & 81.00\%                         & 74.10\%                \\
    + GPT-2 + SBERT          & \textbf{85.25\%}                 & \textbf{82.24\% }                        & \textbf{78.91\% }               \\
    \bottomrule
  \end{tabular}
  \caption{Ablation study on the contribution of JELV2.0's final embedding choices: GPT-2 and SBERT}
  \label{abla}
\end{table}

Table \ref{abla} shows the result of ablation study on the contribution of final embedding choices: GPT-2 and SBERT. This result clearly justifies the inclusion of each embedding component in JELV 2.0.

\section{Meta-evaluation on SEEDA's Test Set of Genetic Domain}
Table \ref{main_table: eval} reports our meta-evaluation on SEEDA’s test set of genetic domain, where our metric still achieves SOTA correlations with human judgments across most evaluation levels and granularities, confirming our metric’s robustness across different domains.

\section{Optimal Hyperparameters}
Table \ref{tab:hyperparams} presents all of the tuned optimal hyperparameters $\alpha$ and $\gamma$ for different evaluation settings across domains.

\section{Statistics about JELV-based Filtering}
\label{app:statistics-JELV}
\begin{table}[ht]
  \centering
  \begin{tabular}{l*{4}{c}}
    \toprule
    \multirow{2}{*}{\textbf{\#Ref. / Sent.}}
      & \multicolumn{2}{c}{\textbf{BEA-Train}}
      & \multicolumn{2}{c}{\textbf{BEA-Dev}} \\
      & Pre-J & Post-J
      & Pre-J & Post-J \\
    \midrule
    Mean      & 11.54 & 3.90  & 14.21 & 4.08  \\
    S.D.      & 4.74  & 3.23  & 6.47  & 3.63  \\
    Max       & 104   & 40    & 78    & 35    \\
    \bottomrule
  \end{tabular}
  \caption{Number of references per sentence in BEA-Train and BEA-Dev before (Pre-J) and after (Post-J) JELV2.0 filtering. Post-J values reflect only edits judged valid, yielding lower averages and reduced variability.}
\end{table}

\section{Complete Prompts}
\subsection{LLM-as-Judges Pipeline (JELV1.0)}
\label{app: JELV1.0}

\paragraph{1st Turn}
Prompt for DeepSeek-V3.

\begin{lstlisting}
You are a linguist tasked with judging whether a proposed sentence edit is VALID or INVALID.
VALID = 1, INVALID = 0.

A VALID edit must satisfy THREE CRITERIA:

  1. GRAMMATICALITY: the resulting SENTENCE is free of grammatical errors.
  
  2. FAITHFULNESS: the edit preserves the INTENDED meaning you infer from the source sentence and its context; acceptable modifications that IMPROVE clarity or expression are allowed as long as the INTENDED meaning is effectively conveyed.
  
  3. FLUENCY: Does the hypothesis edit IMPROVE the fluency of the sentence?

The improvement in fluency can result from enhancements in grammaticality, naturalness, or readability.

ONLY when ALL of the three criteria are satisfied can the edit be judged as valid.

NOTE: Simple synonym substitutions that do NOT measurably IMPROVE clarity or fluency are INVALID.
Refer to the In-Context Examples provided above when making your judgment.

Only output a one-sentence analysis prefixed with ``Analysis'':, then on a new line ``Final Judgment: [0/1]''.
\end{lstlisting}

\paragraph{2nd and 3rd Turn}
Prompt for GPT4o and Qwen.
\begin{lstlisting}
You are an expert language model specializing in evaluating edit judgments within context.

You will receive a JSON object containing:

  'src': the original source sentence.
  'edit': a description of the hypothesis edit.
  'hypo': the sentence after applying the hypothesis edit.
  'llm_analysis': a one-sentence analysis of the edit.
  'llm_prediction': a binary prediction from the previous model (1 = valid, 0 = invalid).

Use the following GOLD CRITERIA to decide if the edit is VALID (1) or INVALID (0):

1. GRAMMATICALITY: the resulting sentence MUST be grammatically correct.

2. FAITHFULNESS: the edit MUST preserve the INTENDED meaning you infer from the source and its context; acceptable modifications that IMPROVE clarity or expression are allowed as long as the INTENDED meaning is effectively conveyed.

3. FLUENCY: Does the hypothesis edit IMPROVE the fluency of the sentence? The improvement in fluency can result from enhancements in grammaticality, naturalness, or readability.

NOTE: Simple synonym substitutions that do NOT contribute to clarity or fluency improvement are NOT considered valid.

ONLY when ALL THREE CRITERIA are fully satisfied should the edit be considered VALID (1).

Your task:
  a) Verify that 'llm_analysis' correctly assesses grammaticality, faithfulness, AND fluency.
  b) Ensure that 'llm_prediction' (0/1) matches that assessment.
If you find any discrepancy, output a revised 'llm_analysis' and adjust the 'llm_prediction'.
Respond with a JSON object containing ONLY two keys:
  'llm_analysis': the corrected one-sentence analysis,
  'llm_prediction': the corrected binary judgment.
\end{lstlisting}

\subsection{Candidate Correction Generation}
\begin{lstlisting}
You are tasked with correcting an English sentence using a reference example. The format is as follows:

Original: {{original_sentence}}
Reference: {{reference_sentence}}

Your goal is to generate ALL POSSIBLE corrections for the original sentence based on the reference sentence, without being restricted by the specific style, modification approach, or pattern of the reference. Please generate corrections that could improve the sentence, ensuring they follow the **GOLD CRITERIA** outlined below.

ONLY generate valid edits, meaning:
- **GRAMMATICALITY**: The correction MUST be free of grammatical errors.

- **FAITHFULNESS**: The correction MUST PRESERVE the **INTENDED meaning** of the original sentence. 
Acceptable modifications that improve clarity or expression are allowed, as long as the intended meaning is preserved.

- **FLUENCY**: The correction MUST IMPROVE the **naturalness** and **readability** of the sentence.

If no valid modifications are possible beyond the reference, output ``ONLY one reference!''. Format the corrections as follows:

[correction 1] ...
[correction 2] ...
[correction N] ...
\end{lstlisting}

\bibliography{aaai2026}

\end{document}